# Using Multimodal Large Language Models for Automated Detection of Traffic Safety Critical Events

Mohammad Abu Tami, Huthaifa I. Ashqar, and Mohammed Elhenawy

*Abstract*—Traditional approaches to safety event analysis in autonomous systems have relied on complex machine learning models and extensive datasets for high accuracy and reliability. However, the advent of Multimodal Large Language Models (MLLMs) offers a novel approach by integrating textual, visual, and audio modalities, thereby providing automated analyses of driving videos. Our framework leverages the reasoning power of MLLMs, directing their output through context-specific prompts to ensure accurate, reliable, and actionable insights for hazard detection. By incorporating models like Gemini-Pro-Vision 1.5 and Llava, our methodology aims to automate the safety critical events and mitigate common issues such as hallucinations in MLLM outputs. Preliminary results demonstrate the framework's potential in zero-shot learning and accurate scenario analysis, though further validation on larger datasets is necessary. Furthermore, more investigations are required to explore the performance enhancements of the proposed framework through few-shot learning and fine-tuned models. This research underscores the significance of MLLMs in advancing the analysis of the naturalistic driving videos by improving safety-critical event detecting and understanding the interaction with complex environments.

## I. INTRODUCTION

The recent advancement breakthrough in Large Language Models (LLMs) has revealed the potential usage in the complex challenging environment of analyzing driving videos. Many researchers investigated the potential of utilizing LLMs in Autonomous Driving through textual representations [1]. With the advancement of multimodal LLM (MLLM), a new merge has been reached with the power reasoning of LLMs in the different modalities of text, image and audio [2], [3].

Autonomous driving is considered one of the complex and critical environments that could benefit from the new MLLM breakthrough. While full autonomous driving might still be far reach. MLLM could advance the understanding of the dynamic variety of road transportation through providing textual analysis of the visual representation of the environment and the different agents in it, then using this analysis to provide a direct, concise, and actionable early warning to the ego-driver in the case of any potential hazards.

Before the era of MLLM, researchers in safety event analysis relied on developing a complex machine learning model from the ground up, utilizing thousands of annotated datasets to achieve high accuracy and reliability. For instance, authors in [4] proposed a supervised encoder-decoder model where a pre-trained ResNet-101 used as encoder to extract the visual and flow features of 17k ego-car dash cam distinct scenarios, and a SAT structure [5] as decoder to predict the caption of the street frames while attained the extracted features from the encoder part.

The study by Chen et al. [6] proposed a pretraining method that aligns numeric vector modalities with LLM (GPT3.5) representations, improving the system's ability to interpret driving scenarios, answer questions, and make decisions. Furthermore, the study titled DriveMLM [7] introduces an LLM-based Autonomous Driving (AD) framework that aligns multi-modal LLMs with behavioral planning states, enabling close-loop autonomous driving in realistic simulators. It bridges the gap between language decisions and vehicle control commands by standardizing decision states according to the off-the-shelf motion planning module.

On another hand, Drive As you Speak paper [8] presents an approach to enabling human-like interaction with large language models in autonomous vehicles. It leverages LLMs to understand and respond to human commands, demonstrating the potential of LLMs in creating more intuitive and user-friendly autonomous driving experiences.

Despite the promising developments in utilizing Large Language Models (LLMs) for autonomous driving, a significant gap remains in the application of these models for safety-critical event analysis. Existing research focuses on enhancing autonomous driving capabilities through improved perception and decision-making processes without specifically addressing the unique challenges posed by safety-critical situations. This gap highlights the need for a specialized approach that leverages the multimodal capabilities of LLMs to directly address the nuances of safety-critical events in driving scenarios.

Our contribution aims to bridge this gap by introducing a Multimodal Large Language Model (MLLM) framework specifically designed for the analysis and interpretation of safety-critical events. By integrating the diverse modalities of text and image, our framework seeks to provide a more holistic and nuanced understanding of dynamic driving environments. Furthermore, our approach emphasizes the automation of extracting visual representation from the raw video and fed it to MLLM with a creation of context-specific prompts to guide the MLLM's analysis, focusing on generating actionable insights for hazard detection and response. This research presents a novel application of MLLMs in a domain where

M. Abu Tami is with Arab American University, Jenin, Palestine, (E-mail: m.abutami@student.aaup.edu).
H. I. Ashqar is with Arab American University Jenin, Palestine, and Columbia University, NY, USA (E-mail: huthaifa.ashqar@aaup.edu).
M. Elhenawy is with CARRS-Q, Queensland University of Technology Brisbane, Australia, (E-mail: mohammed.elhenawy@qut.edu.au).

precision and reliable decision-making are essential, marking a significant step forward in the pursuit of safer driving.

## II. METHODOLOGY

### A. Dataset

Creating a dataset from driving videos that integrates language for visual understanding is a challenging task. This process is a resource-extensive task that requires trained human annotators for optimal accuracy and reliability. In addition, the variety and complexity of driving scenarios require a dataset rich in visual scenes. The dataset needs to cover a variety ranging from simple driving directions to complex situations involving pedestrians, other vehicles, and road signs.

Many researchers have either enhanced existing datasets with textual information [9],[10],[11], [12] or developed new ones from scratch [4],[13]. Notable among these are the DRAMA datasets [4]. DRAMA, in particular, focuses on driving hazards and related objects, featuring video and object-level inquiries. This dataset supports visual captioning with free-form language descriptions and accommodates both closed and open-ended questions, making it essential for assessing various visual captioning skills in driving contexts. In addition, the vast variety found in DRAMA which consists of ~17K distinct scenarios make it a uniquely comprehensive resource for investigating and evaluating MLLM models on complex driving situations.

Considering these factors, this study has selected the DRAMA dataset for its and utilizing the ground truth labelled gathered from DR experiments AMA to report this paper's experiment results. DRAMA's detailed focus on hazard detection and its comprehensive framework for handling natural language queries make it exceptionally suitable for pushing forward research in safety critical event analysis.

### B. Proposed Framework

To establish a deterministic and safe MLLM-based framework for the safety-critical event analysis, the work focuses on implementing a controlled model response without sacrificing the free-form power of the LLM output. Given the hallucinating problem that affects the LLMs [14], the research methodology involves experiments with two foundational MLLMs: the Gemin-pro-vision developed by Google [15] and the Large Language and Vision Assistant (Llava) model [16]. By integrating these models with the presented framework, authors seek to harness their computational power while mitigating risks associated with their output, ensuring more reliable and contextually accurate responses.

The framework illustrated in Fig. 1 is designed for detecting safety-critical events in from driving video extracted from car dash cam, utilizing a multi-stage Question-Answer (Q/A) approach with a multi-modal large language model (MLLM) [4]. The process initiates with "Frame Extraction," where the system automatically collects video frames from the ego vehicle's camera at regular intervals (every second). These frames are subjected to a "Hazard Detection" phase, where the model assesses the scene for potential dangers.

Upon identifying a hazard, the framework employs a tripartite categorization strategy to probe the nature of the threat further, using "What," "Which," and "Where" queries to reveal the object-level details. In the "What" phase, the MLLM classifies the entities detected by the camera, differentiating among agents like pedestrians, vehicles, or infrastructure elements. The "Which" stage involves the MLLM identifying specific features and attributes of these agents, such as pedestrian appearance, vehicle make and model, or infrastructure type, providing vital contextual insights for decision-making.

The final "Where" phase tasks the MLLM with determining the spatial location and distance of the hazard agents from the ego-car, including their position on the road, proximity to the vehicle, and movement direction. This spatial information is essential for the ego-car system $t$ Automated to make safe navigation decisions.

### C. Video Analysis Strategies

To enhance the efficiency of the proposed framework, two distinct methodologies were experimented for analyzing the volume of vehicle video data. These methodologies enable the system to focus on the most relevant information, thereby optimizing processing speed and accuracy in detecting safety-critical events.

#### 1) Sliding Window Frame Capture

The first methodology employs a sliding window approach to systematically capture and analyze subsets of video frames. This technique involves defining a window that slides over the video timeline, capturing a specific range of frames from $t_i$ to $t_{i+n}$ where $t_i$ represents the initial frame in the window, and n is a configurable variable that determines the number of frames included in each window. The mathematical representation of this method can be expressed as follows:

$$\text{Window}(t_i) = \{\text{Frame}(t_i), \text{Frame}(t_{i+1}), \dots, \text{Frame}(t_{i+n-1})\} \quad (1)$$

This strategy allows for the dynamic adjustment of the window size based on the specific requirements of the analysis, enabling the system to balance data comprehensiveness and processing efficiency.

#### 2) Textual Context Representation

The second methodology integrates MLLM to convert past visual frames into a textual context representation. This approach leverages the MLLM's capability to succinctly describe visual data in text form, encapsulating the essential elements and their interactions within the frame. Once a textual context representation is generated, it is combined with a sequence of frames captured using the sliding window method described above. The combined data then serves as input for predicting safety-critical events. The process can be mathematically represented as:

$$\text{Context}(t_i) = \text{MLLM}(\text{Frame}(t_{i-n}), \dots, \text{Frame}(t_{i-1})) \quad (2)$$
$$\text{Prediction} = \text{MLLM}(\text{Context}(t_i), \text{Window}(t_i)) \quad (3)$$

where Context($t_i$) is the textual representation of the frames from $t_{i-n}$ to $t_{i-1}$ as generated by MLLM and Window($t_i$) includes the frames from $t_i$ to $t_{i+n}$. The Prediction in equation (3) is the output of combining these inputs into MLLM.

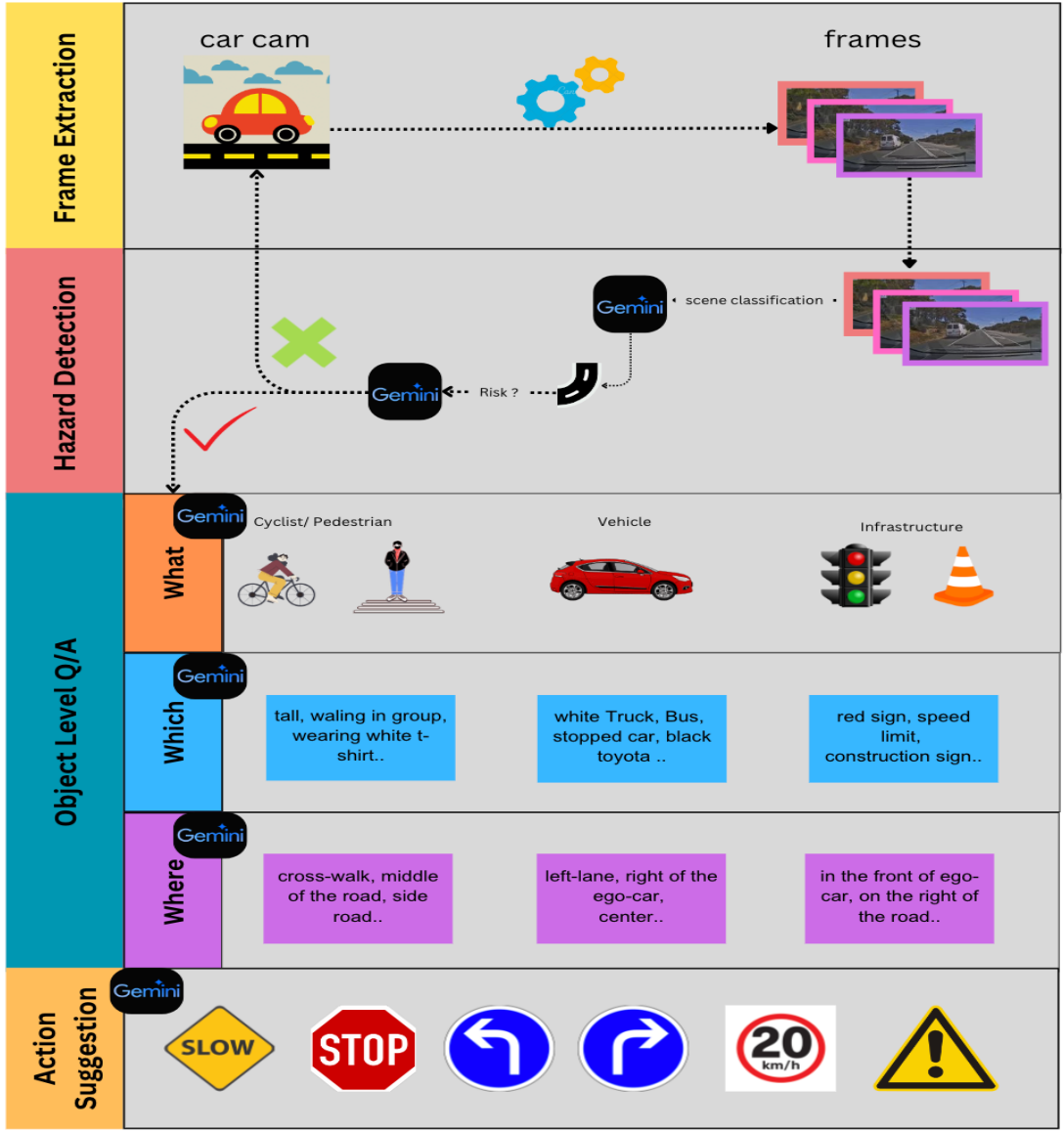

Figure 1. Automated Multi-Stage Hazard Detection Framework for Safety Critical Event Using MLLM.

### D. Data Augmentation Strategy

Various image augmentation techniques were applied to the images before they were fed into the MLLM for hazard-detection question answering (QA). These augmented images aim to direct the MLLM to different areas within the language distribution it relies on for generating responses as illustrated in Fig. 2. The key idea of using different augmentation for the same scene under investigation is to direct the model to different places in the language distribution, which could help the model of generating more textual representation of the scene when generating response through local sampling.

By introducing augmented images in the prompt, MLLM can start at various points within the data distribution, which influences the diversity of local sampling results. Subsequently, the outcomes from different model sampling processes are aggregated using a top-k voting mechanism to determine the outcome response. This approach aims to aid the model in producing textual responses that more accurately represent the scene under querying.

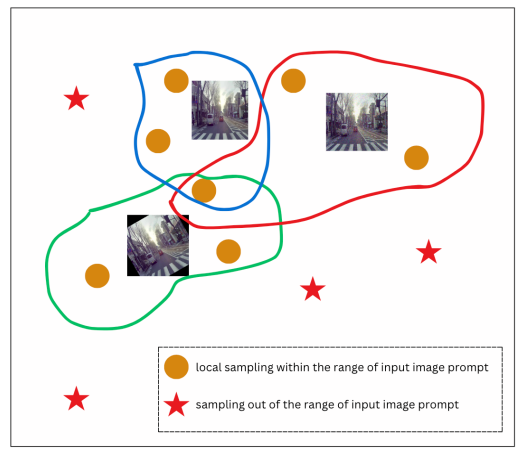

Figure 2. A Conceptual 2-D Diagram of Augmented Image Prompting.

## III. RESULTS

The work presented in this paper demonstrates the potential of leveraging the capabilities of MLLM in analyzing safety-critical event scenarios using multi-modal data integration and dynamic contextual data reduction for guiding the model's output. The prediction illustrated in Fig. 3 showcases the proficiency of Gemini-pro-vision 1.5 in zero-shot learning scenarios.

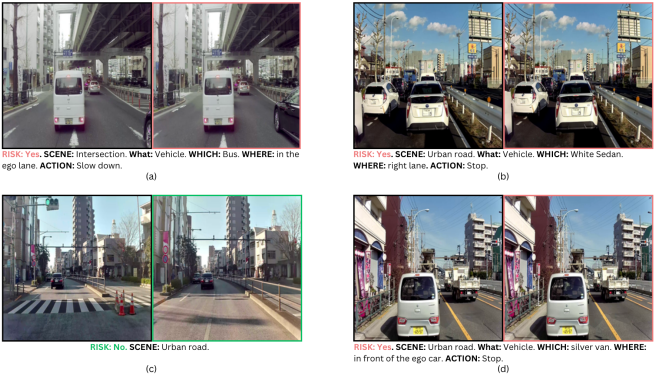

Figure 3. Outputs from Gemini-Pro-Vision 1.5, Analysis with Sliding Window ($n = 2$).

To conclude the effectiveness of the proposed framework, a series of experiments were carried out utilizing two distinct base models: Gemini-pro-vision 1.5 by Google and Llava-7B 1.5. The findings, as detailed in TABLE I, and visually presented in Fig. 4 underscore the exceptional capabilities of Gemini-pro-vision 1.5 in various question-answering stages. Gemini-pro-vision's outstanding performance can be attributed to its larger size and the extensive dataset it was trained on, providing it with a broader understanding and a more nuanced ability to interpret and answer questions across different stages. On the other hand, Llava-7B 1.5 has shown notable potential, especially when considering its results in certain Q/A stages. While its overall scores were lower, it's probably with fine-tuning datasets more closely related to driving scenarios, Llava-7B 1.5's effectiveness could improve significantly.

TABLE I. COMPARATIVE PERFORMANCE ANALYSIS OF Q/A FRAMEWORKS USING SLIDING WINDOW REDUCTION AND TEXTUAL CONTEXT REPRESENTATION.

| Q/A | Sliding Window Reduction ($n = 2$) | | Textual Context Representation | | Video |
|---|---|---|---|---|---|
| | *Gemini-pro-vision 1.5* | *Llava-7B 1.5* | *Gemini-pro-vision 1.5* | *Llava-7B 1.5* | *Gemini-pro-video* |
| *Risk* | 60% | 55% | 55% | 45% | 75% |
| *Scene* | 95% | 80% | 90% | 75% | 100% |
| *What* | 90% | 75% | 85% | 70% | 65% |
| *Which* | 95% | 60% | 85% | 60% | 65% |
| *Where* | 80% | 60% | 60% | 50% | 65% |
| *Proposed Action* | 70% | 55% | 55% | 50% | 75% |
| **Overall** | **81.6%** | **64.1%** | **71.6%** | **58.3%** | **74.1%** |

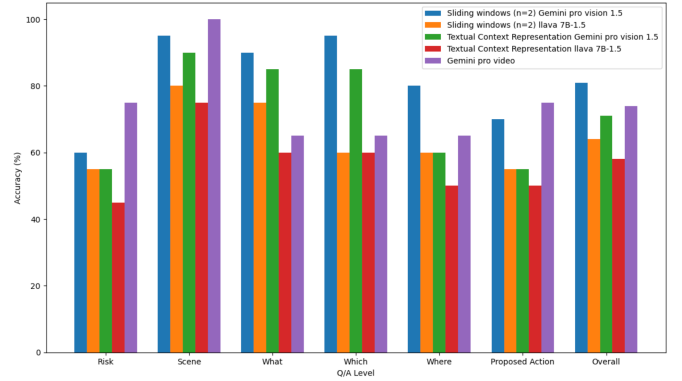

Figure 4. Comparative Analysis of Methodology Performance Across Seven Q/A levels.

The result from the augmented images experiment revealed that the best result accuracy is when no augmentation was applied to the image as seen in TABLE II and visualized in Fig. 5. The Q/A frameworks are tested on raw images as well as images subjected to rotation by 30 degrees and noise augmentation. Additionally, the top-k voting method is applied to aggregate predictions from multiple models. The results indicate varying levels of performance across different augmentation techniques, with raw images generally outperforming rotated and noisy images. However, the top-k voting approach yields improved accuracy compared to individual models, suggesting its effectiveness in enhancing the overall performance of the Q/A frameworks.

The fact that augmentation causes a degradation in the result accuracy might be to the fact that the model was not able to determine the scene and agent when the augmented image applied. However, it is worth mentioning that further investigation might reveal different results if applying the augmentation to more samples and using a wide range of augmentation techniques. This analysis provides insights into the impact of image augmentation and ensemble techniques on the accuracy and robustness of Q/A systems, facilitating informed decisions for model selection and deployment in practical applications.

TABLE II. COMPARATIVE PERFORMANCE ANALYSIS OF Q/A FRAMEWORKS USING AUGMENTED IMAGES AND TOP-K VOTING.

| Q/A | Raw | Rotate 30 | Noise | Top-k voting |
|---|---|---|---|---|
| *Risk* | 60% | 55% | 40% | 55% |
| *Scene* | 95% | 85% | 70% | 90% |
| *What* | 90% | 40% | 10% | 55% |
| *Which* | 95% | 25% | 0% | 35% |
| *Where* | 80% | 15% | 5% | 30% |
| *Proposed Action* | 70% | 40% | 30% | 45% |
| **Overall** | **81.6%** | **43.3%** | **25%** | **51.6%** |

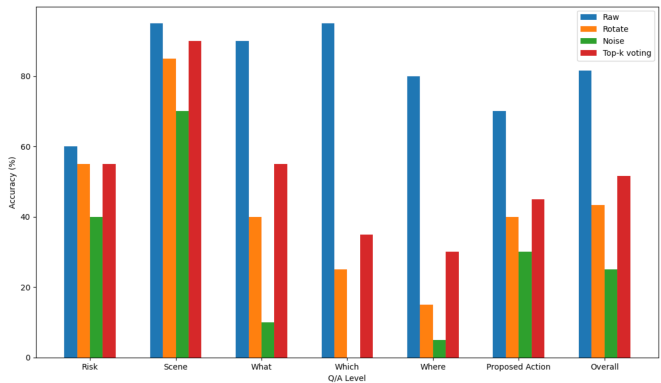

Figure 5. Performance Evaluation of Augmentation Methodologies.

## IV. Discussion

The results of the comparative analysis highlight the critical role of data augmentation techniques in enhancing the robustness of Q/A frameworks. While augmented images may introduce noise and distortions, they also provide valuable training data that can improve the model's ability to generalize to unseen scenarios. However, our findings suggest that caution should be exercised when applying augmentation methods, as they may not always lead to performance improvements.

In real-world applications, the choice of augmentation techniques should be guided by the specific characteristics of the target dataset and the requirements of the Q/A system. For example, in scenarios where the training dataset lacks diversity or representative samples, augmentation methods such as rotation and noise augmentation can help mitigate overfitting and improve model generalization. Conversely, in datasets with sufficient variability, the use of raw images may suffice, as they preserve the integrity of the original data without introducing additional complexity.

Furthermore, the adoption of ensemble methods such as top-k voting holds significant promise for enhancing the reliability and robustness of Q/A systems in real-world deployment. By leveraging the collective intelligence of multiple models, ensemble techniques can mitigate individual model biases and uncertainties, leading to more accurate and trustworthy predictions. This has profound implications for applications in safety-critical domains such as autonomous driving, medical diagnosis, and industrial automation, where the reliability of decision-making systems is paramount.

## V. Conclusion

Our contribution aims to address the gap in safety-critical event analysis and interpretation by introducing a MLLM framework. This framework is specifically designed to integrate text and image modalities, providing a comprehensive understanding of dynamic driving environments. By automating the extraction of visual representations from raw video data and feeding it into the MLLM alongside context-specific prompts, our approach focuses on generating actionable insights for hazard detection and response. This novel application of MLLMs in the domain of driving safety represents a significant advancement, emphasizing precision and reliable decision-making in critical situations. Our framework has the potential to contribute to safer driving practices by enabling a deeper understanding of complex driving scenarios and facilitating proactive risk mitigation strategies.

The results noticed when experiment with sliding windows and textual-context representation has shown significant potential in preliminary tests. On another hand, while augmented images with top-k voting does not reveal an improvement over the raw image result, further investigation with more variety of augmented technique could reveal different results.

The comparative performance analysis of question-answering (Q/A) frameworks using augmented images and top-k voting techniques reveals valuable insights into their effectiveness and robustness. Across various augmentation methods, including rotation by 30 degrees and noise augmentation, raw images consistently outperform their augmented counterparts in terms of accuracy for risk detection, scene comprehension, object identification, object classification, location determination, and proposed action determination. However, the top-k voting method demonstrates its potential to enhance overall performance by aggregating predictions from multiple models. These findings underscore the importance of careful consideration when selecting augmentation techniques and ensemble methods for Q/A frameworks, balancing the trade-offs between accuracy and computational efficiency.

Future research will focus on expanding the result of the proposed technique through employing a rich set of configurations, this includes studying the impact of sliding windows size on the model response. Furthermore, authors will investigate more MLLM models and apply more image analysis techniques to establish more reliable and robust results.